\documentclass[10pt, a4paper]{article}

\usepackage{lrec-coling2024} 
\usepackage{amsmath, bm}
\usepackage{amssymb}
\usepackage{array}
\usepackage{mathrsfs}
\usepackage{makecell}
\usepackage{booktabs}
\usepackage{multirow}
\title{\text{L}$^{2}$GC: Lorentzian Linear Graph Convolutional Networks For Node Classification\\ \vspace*{.5\baselineskip} }

\name{Qiuyu Liang$^1$, Weihua Wang$^{1,2,3\;*}$ \thanks{*Corresponding author.}, Feilong Bao$^{1,2,3}$, Guanglai Gao$^{1,2,3}$} 

\address{$^1$ College of Computer Science, Inner Mongolia University, Hohhot, China \\
         $^2$ National and Local Joint Engineering Research Center of Intelligent Information Processing \\ Technology for Mongolian, Hohhot, China \\
         $^3$ Inner Mongolia Key Laboratory of Mongolian Information Processing Technology, Hohhot, China \\
         liangqiuyu@mail.imu.edu.cn, wangwh@imu.edu.cn\\
         }

\abstract{
Linear Graph Convolutional Networks (GCNs) are used to classify the node in the graph data. However, we note that most existing linear GCN models perform neural network operations in Euclidean space, which do not explicitly capture the tree-like hierarchical structure exhibited in real-world datasets that modeled as graphs. In this paper, we attempt to introduce hyperbolic space into linear GCN and propose a novel framework for Lorentzian linear GCN. Specifically, we map the learned features of graph nodes into hyperbolic space, and then perform a Lorentzian linear feature transformation to capture the underlying tree-like structure of data. Experimental results on standard citation networks datasets with semi-supervised learning show that our approach yields new state-of-the-art results of accuracy 74.7$\%$ on Citeseer and 81.3$\%$ on PubMed datasets. Furthermore, we observe that our approach can be trained up to two orders of magnitude faster than other nonlinear GCN models on PubMed dataset. Our code is publicly available at \href{https://github.com/llqy123/LLGC-master}{https://github.com/llqy123/LLGC-master}.
 \\ \newline  
 }

\begin{document}

\maketitleabstract

\section{Introduction}

We consider the problem of classifying nodes (such as documents) on the graph-structured data (such as citation network), where models need to learn information from their neighbours.
Graph Convolutional Network (GCN) is one of classification models that be able to aggregate and propagate the node information in graphs.
Therefore, GCNs and their variants have achieved success in various tasks, such as text classification \citep{yao2019graph}, relation extraction \citep{tian2021dependency}, recommendation systems \citep{ying2018graph,shang2019gamenet, wang2021self, feng2022graph} and social networks analysis \citep{li2019encoding, sankar2021graph}.

In order to learn the features of each node, GCNs typically consist of two successive stages: node feature propagation and transformation. 
In the first stage, GCNs stacked multiple layers by a nonlinear activation function to aggregate and utilize nodes information from neighboring nodes.
This multi-layer structure can obtain both local and global features naturally, yet the computation complexity increased rapidly and further hinder their applications.
Therefore, some researchers attempted to improve the GCNs by eliminating the nonlinear active function between layers, such as \citet{wu2019simplifying,wang2021dissecting,li2022g}, which are referred to as linear GCN model.
\begin{figure}[]
\begin{center}
\includegraphics[scale=0.41]{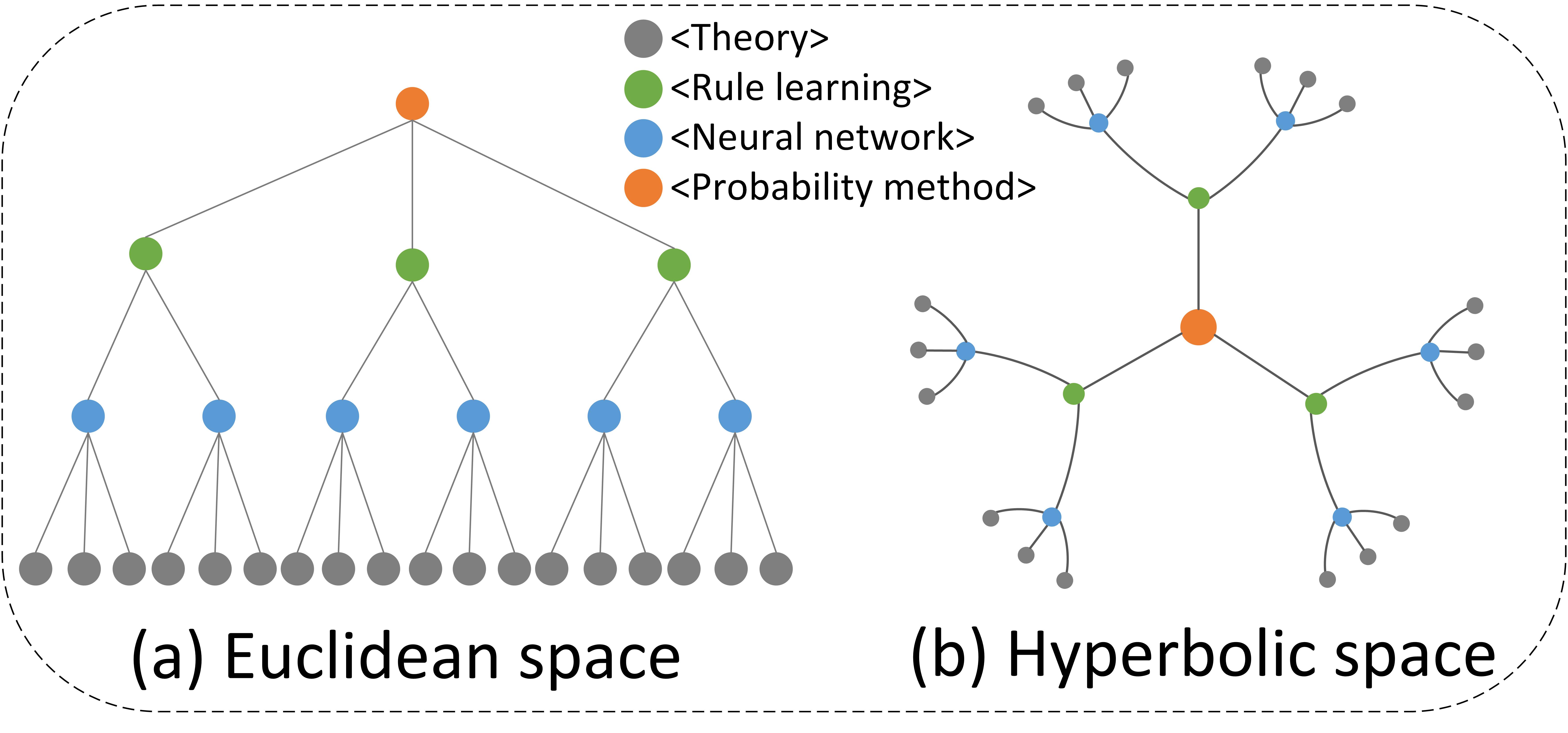} 
\caption{Visualization of a tree-like hierarchical structure citation network in Euclidean and hyperbolic space. In both graphs, nodes represent documents, and edges represent citations. Each color node represents a type of document.}
\label{fig1}
\end{center}
\end{figure}
However, these traditional linear GCN models mainly focus on improving the nodes feature propagation scheme, while pay fewer attention to the feature transformation stage. 

In the second stage, GCNs will transform the learned features and structural information to further classify nodes.
The feature transformation is usually performed in Euclidean space.
In fact, real-world datasets typically exhibit tree-like hierarchical or scale-free structures, but feature transformation in Euclidean space may be distorted when faced with this kind of data.
Unlike Euclidean spaces, hyperbolic spaces have a greater capacity to learn scale-free or hierarchical structures graph, which can be used to learn tree-like graphs~\citep{clauset2008hierarchical, muscoloni2017machine}. 
For example, we illustrated the same tree-like graph instance within Euclidean and hyperbolic spaces in Figure~\ref{fig1}.
From Figure~\ref{fig1}, we obvious that the distance between two nodes will be larger when mapping to the hyperbolic space.
Since the distance between two nodes is the Euclidean distance, while it will change to a geodesic distance in hyperbolic space.
Furthermore, the local dependencies and hierarchical relationships between nodes can be preserved simultaneously when mapping into hyperbolic space.

To address the limitation of feature transformation in Euclidean space,
we propose a novel \textbf{L}orentzian \textbf{L}inear \textbf{G}raph \textbf{C}onvolutional Networks (\textbf{$\text{L}^{2}$GC}) for classifying the tree-like graph node. 
In contrast to its Euclidean space counterparts, we modeled the node feature transformation with Lorentzian manifold. 
The main contributions of this study can be summarized as follows: 

\begin{itemize}
    \item {To the best of our knowledge, our work is the first to present a framework for modeling the feature transformation of a linear GCN model in hyperbolic space with Lorentzian manifold.}
    \item {Our approach can be trained two orders of magnitude speedup over nonlinear models on the large-scale dataset (Pubmed) in our evaluation while be computationally more efficient and fitting significantly fewer parameters.}
    \item {Extensive experiments on semi-supervised node classification tasks show that our model achieves state-of-the-art results on tree-like datasets compared with other nonlinear or linear GCN models in different spaces.}
\end{itemize}

\section{Related Work}
\subsection{Linear Graph Convolutional Networks}
Traditional GCNs are commonly used to process graph data since their abilities to aggregate and propagate information among nodes, but these models often burden excess complexity that inherit from their neural networks lineage. 
Recent efforts try to improve linear GCNs on the processing of node information propagation. 
For example, \citet{wu2019simplifying} simplified GCNs by repeatedly removing the nonlinearities between GCN layers and folding weight matrices between layers to form the new objective function into a single linear transformation. 
\citet{wang2021dissecting} introduced Decoupled Graph Convolution which decouples the terminal time and the feature propagation steps. 
\citet{li2022g} introduced a novel propagation layer from a spectrum perspective. 
This new layer decoupled the three concentration properties: concentration center, maximum response and bandwidth, which implement a low-pass graph filter for a linear graph model.
In addition, fully linear GCNs are proposed in~\citet{cai2023fully}.
This is a fully linear GCN that both straightforward and efficient for semi-supervised and unsupervised learning tasks.

While above linear GCNs have achieved competitive performance, they only focus on the graph node feature propagation stage and overlook the feature transformation stage. 
They perform feature transformation operations in Euclidean space, which leads to greater distortion when faced with real-world graphs with scale-free or hierarchical structures. 
Different from the above work, our work takes the foundation of linear GCNs and transforms the feature within hyperbolic spaces.

\subsection{Hyperbolic Graph Convolutional Networks}
Recently, a growing number of researchers have investigated hyperbolic geometry since their greater capacity to learn graphs that characterized by scale-free or hierarchical structures.
For example, \citet{chami2019hyperbolic} lifted graph convolutional network to hyperbolic geometry and then proposed hyperbolic Graph Convolutional Network (HGCN) model. 
\citet{zhang2021hyperbolic} studied the Graph Neural Network with attention mechanism in hyperbolic spaces at the first attempt. 
To address the issue of the HGCN model's operation on the tangent space that dissatisfied definition of hyperbolic geometry, \citet{zhang2021lorentzian} introduced the Lorentzian version of traditional GCN and reconstructed the graph operations of hyperbolic GCN.
\citet{chen2022fully} devised a completely hyperbolic framework without tangent spaces to address the limitation of hyperbolic GCN models. 

Although existing hyperbolic GCN models could address the challenge of scale-free or hierarchical structures when handling graph, these models often suffer from complex structures and large amount of parameters.
In order to reduce the time cost and number of parameters, we try to simplify the calculation process in hyperbolic space.

\section{Preliminaries}
In this section, we briefly review the key concepts of hyperbolic geometry. 
A thorough and detailed explanation of hyperbolic geometry can be found in~\citep{willmore2013introduction}. 
Hyperbolic geometry is a non-Euclidean geometry with constant negative curvature $k\;(k>0)$.
Several hyperbolic geometric models have been utilized in previous studies, including the Poincaré ball model \citep{ganea2018hyperbolic}, the Poincaré half-plane model \citep{tifrea2018poincar}, the Klein model \citep{gulcehre2018hyperbolic}, and the Lorentz model \citep{nickel2018learning}. 
These models are mathematically equivalent.

We choose the Lorentz model as the eigenspace in our framework due to the numerical stability and computational simplicity provided by its exponential/logarithmic maps.

\subsection{The Lorentz Model}
An $n$-dimensional Lorentz model is the Riemannian manifold $\mathbb{L}_{k}^{n} = (\mathcal{L}^{n},\mathfrak{g}^{k}_{X})$ with negative curvature $k\;(k>0)$.
$\mathfrak{g}^{k}_{X} = diag\left ( -1,1,\cdots ,1\right )$ is the Riemannian metric tensor.
$\mathcal{L}^{n}=\left \{\texttt{x}\in \mathbb{R}^{n+1}|\left \langle \texttt{x},\texttt{x}\right \rangle_{\mathcal{L}}=\frac{1}{k},x_{0}>0 \right \}$ is a point set, which corresponds to the upper sheet of a hyperboloid in an ($n+1$) dimensional Minkowski space with the origin.
$\left \langle \texttt{x},\texttt{y} \right \rangle_{\mathcal{L}}$ is the Lorentzian inner product of two point \texttt{x} and \texttt{y} in Riemannian manifold, which is defined as follows:
\begin{equation}
\label{equation5}
\left \langle \texttt{x},\texttt{y} \right \rangle_{\mathcal{L}}=-x_0y_0 + \sum_{i=1}^{n}x_iy_i,
\end{equation}

\noindent
\textbf{Tangent Space.} 
Each point in Riemannian manifold $\mathbb{L}_{k}^{n}$ has the form of $\texttt{x} = \begin{bmatrix}x_{0} \\ x_{s} \end{bmatrix}$, where $\texttt{x} \in \mathbb{R}^{n+1}, x_0 \in \mathbb{R}, \texttt{x}_s \in \mathbb{R}^{n}$. 
Let $\texttt{x}$ denote as a point in $\mathbb{L}_{k}^{n}$. 
Considering the Lorentzian inner product (Eq. \ref{equation5}), the orthogonal space of $\mathbb{L}_{k}^{n}$ at $\texttt{x}$ can be represented as:

\begin{equation}
\label{equation3}
    \mathcal{T} _\texttt{x}\mathbb{L}_{k}^{n}=\left \{\texttt{y}\in \mathbb{R}^{n+1}|\left \langle \texttt{y},\texttt{x}\right \rangle_{\mathcal{L}^{n}}=0 \right \},
\end{equation}
where $\mathcal{T} _\texttt{x}\mathbb{L}_{k}^{n}$ is a Euclidean subspace of $\mathbb{R}^{n+1}$.
Specifically, the tangent space at the origin is defined as $\mathcal{T} _\texttt{0}\mathbb{L}_{k}^{n}$.

\noindent
\textbf{Exponential and Logarithmic Maps.} 
Given a point $\texttt{x}\in \mathcal{T} _\texttt{x}\mathbb{L}_{k}^{n}$ in tangent space.
The tangent space is a Euclidean subspace.
The exponential map is a mapping from the tangent space $\mathcal{T} _\texttt{x}\mathbb{L}_{k}^{n}$ to the hyperbolic space $\mathbb{L}_{k}^{n}$, which move along the geodesic $\gamma$, satisfying $\gamma\left(0\right)=\texttt{x}$ and ${\gamma}'\left(0\right)=\texttt{z}$.
The exponential map can be defined as:
\begin{equation}
\begin{aligned}
\label{equation1}
    &\text{exp}_{\texttt{x}}^{k}\left ( \texttt{z} \right )=\text{cosh}\left ( \alpha \right )\texttt{x} + \text{sinh}\left ( \alpha \right )\frac{\texttt{z}}{\alpha },\\
    &\alpha =\sqrt{-K}\left \| \texttt{z}\right \|_{\mathcal{L} },\left \| \texttt{z}\right \|_{\mathcal{L} }=\sqrt{\left \langle \texttt{z},\texttt{z}\right \rangle}_{\mathcal{L}},
\end{aligned}
\end{equation}

The logarithmic map is the reversed mapping that maps back to the tangent space $\mathcal{T} _\texttt{x}\mathbb{L}_{k}^{n}$.
Given a point $\texttt{y}\in \mathbb{L}_{k}^{n}$ in hyperbolic space, the logarithmic map can be defined as:
\begin{equation}
\begin{aligned}
\label{equation2}
    \text{log}_{\texttt{x}}^{k}\left ( \texttt{y}\right )&=\frac{\text{cosh}^{-1}\left ( \beta \right )}{\sqrt{\beta ^{2}-1}}\left ( \texttt{y}-\beta \texttt{x}\right ),\\
    \beta &=k\left \langle \texttt{x},\texttt{y}\right \rangle_{\mathcal{L} }
\end{aligned}
\end{equation}

\begin{figure*}[h!]
\begin{center}
\includegraphics[width=1\linewidth]{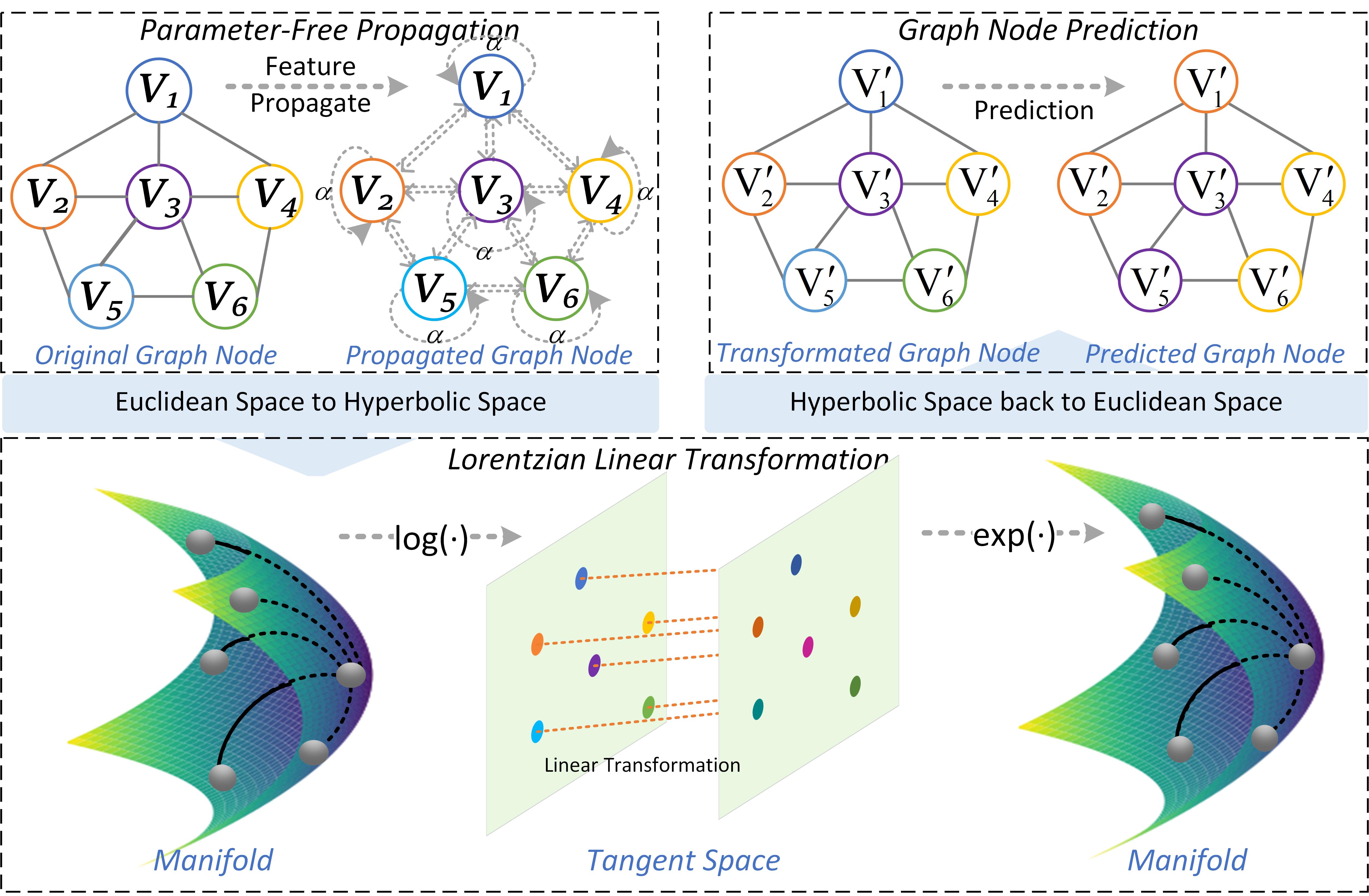} 
\caption{The framework of Lorentzian Linear Graph Convolutional Networks for node Classification.}
\label{fig2}
\end{center}
\end{figure*}

\subsection{Lorentzian version of Function}
The Lorentzian version of a function $f^{\otimes}$ to map from $\mathbb{L}_{k}^{n}$ to $\mathbb{L}_{k}^{m}$ for two points $\texttt{x}=\left ( x_{0},\cdots ,x_{n}\right )\in \mathbb{L}_{k}^{n}$ and $\texttt{v}=\left ( v_{0},\cdots ,v_{n}\right )\in \mathcal{T} _\textbf{0}\mathbb{L}_{k}^{n}$ is defined as:
\begin{equation}
\begin{aligned}
    f^{\otimes^{k}}\left ( \texttt{x}\right )&=\text{exp}_{0}^{k}\left ( \hat{f}\left (\text{log}_{0}^{k}\left ( \texttt{x}\right )\right )\right ),\\
    \hat{f}(\texttt{v})&=\left ( 0,f\left ( v_{1},\cdots ,v_{n}\right )\right ),
\end{aligned}
\end{equation}
where $f:$ $\mathbb{R}^{n}\rightarrow\mathbb{R}^{m}\left ( n,m > 2\right),$ $\text{exp}_{0}^{k}: \mathcal{T} _{0}\mathbb{L}_{k}^{n}\rightarrow\mathbb{L}_{k}^{m}$, $\text{log}_{0}^{k}: \mathbb{L}_{k}^{n}\rightarrow\mathcal{T} _{0}\mathbb{L}_{k}^{m}$. 
$k$ is the constant curvature.

Given a point $\texttt{v}=\left ( v_{0},\cdots ,v_{n}\right )\in \mathcal{T} _{0}\mathbb{L}_{k}^{n}$, existing methods directly apply Euclidean transformations to all coordinates $\left ( v_{0},\cdots ,v_{n}\right )$ within tangent spaces, such as~\citep{chami2019hyperbolic, liu2019hyperbolic}. 
Different from previous methods, canonical Lorentzian version transformation only leverages the Euclidean transformations on the last $n$ coordinates$\left ( v_{1},\cdots ,v_{n}\right )$ in tangent spaces, and the first coordinate $\left ( v_{0}\right )$ is set to $0$, which satisfying the constraint in Eq. (\ref{equation3}).

\section{Methodology}
Here, we describe our proposed framework approach, which is depicted in Figure \ref{fig2}. 
Our framework mainly consists of three steps: parameter-free neighborhood feature propagation in Euclidean space, Lorentzian linear feature transformation in hyperbolic space and graph node labels prediction in Euclidean space.
According to the paper \citep{wu2019simplifying}, it has been observed that the inclusion of a nonlinear activation layer can impact the performance of linear GCNs. 
Therefore, in our framework, we remove the nonlinear activation layer after the Lorentzain feature transformation layer to simplify the model architecture and further improve the performance of our model. 

\subsection{Parameter-Free Propagation}
To mitigate the over-smoothing problem, we adopt a personalized propagation scheme to effectively balance the contributions of graph structure and node features. 
As shown in Figure \ref{fig2}, given a graph $G=\left(\mathcal{V},E\right)$ with $n$ nodes and $m$ edges, where $\mathcal{V}$ represents the set of nodes and $E$ represents the set of edges. 

In addition, $\tilde{G}=\left(\mathcal{V},\tilde{E}\right)$ is defined as the self-looped graph, we attach a self-loop to each node in the original graph $G$. 
Let denote the adjacency matrix of $G$ as \texttt{A} and the diagonal degree matrix as \texttt{D}. 
Therefore, we also define the adjacency matrix and diagonal degree matrix of the self-looped graph $\tilde{G}$ as $\tilde{\texttt{A}}=\texttt{A}+\texttt{I}$ and $\tilde{\texttt{D}}=\texttt{D}+\texttt{I}$, where $\texttt{I}$ is the identity matrix,
$\texttt{X}\in\mathbb{R}^{n\times d}$ is the node feature matrix, and each node $v$ is associated with a $d$ dimensional feature vector represented by $\texttt{X}_{v}$. 

The linear propagation matrix \texttt{P} can be written as: 
\begin{equation}
\begin{aligned}
\texttt{P}=\left(\texttt{D}+\texttt{I}\right)^{-1/2}\left(\texttt{A}+\texttt{I}\right)\left(\texttt{D}+\texttt{I}\right)^{-1/2}.
\end{aligned}
\end{equation}
We precompute the information transfer between a node and its $n$-power neighbors as follows:
\begin{equation}
\label{equation6}
    \texttt{H}^{\left (l+1\right )}=\left ( 1-\alpha \right )\texttt{P}\texttt{H}^{\left (l\right )}+\alpha\texttt{X},
\end{equation}
where $\alpha\in \left (0,1 \right ]$ is the teleport probability of the topic-sensitive PageRank.
The initial node features $\texttt{X}$ as topics to be ranked in the topic-sensitive PageRank algorithm.
The nodes feature propagation with Eq. (\ref{equation6}) guarantees that $\texttt{H}^{\left(l\right)}$ is 
consistently influenced by both the graph structure and the initial node features $\texttt{X}$ with a fixed proportion $\alpha$, which enables us to acquire the graph node features $\texttt{H}^{\left ( n\right )}$ obtained through $n$-power propagations.

\subsection{Lorentzian Linear Transformation}
Theoretically, the linear transformation operations in Euclidean space cannot be directly applied in the hyperbolic space. 
Consequently, we define the linear transformation under Lorentz model to ensure the features transformation processing adhere to the hyperbolic geometry.
In order to apply the linear transformation in hyperbolic space, following Lorentz version, we derive the Lorentzian matrix-vector multiplication as:

\textbf{Definition:} Given two points $\texttt{x}=\left ( x_{0},\cdots ,x_{n}\right )\in \mathbb{L}_{k}^{n}$,  $\texttt{v}=\left ( v_{0},\cdots ,v_{n}\right )\in \mathcal{T} _{0}\mathbb{L}_{k}^{n}$, 
and $\textrm{M}$ is a linear map from $\mathbb{R}^{n}\rightarrow\mathbb{R}^{m}$ $ \left ( n,m > 2\right)$ with matrix representation, we have:
\begin{equation}
\begin{aligned}
\label{equation4}
    \textrm{M}^{\otimes^{k}}\left ( \texttt{x}\right )&=\text{exp}_{0}^{k}\left ( \hat{\textrm{M}}\left (\text{log}_{0}^{k}\left ( \texttt{x}\right )\right )\right ),\\
    \hat{\textrm{M}}(\texttt{v})&=\left ( 0,\textrm{M}\left ( v_{1},\cdots ,v_{n}\right )\right ).
\end{aligned}
\end{equation}
$\textrm{M}$ is a $m\times n$ matrix, $\textrm{M}'$ is a $l\times n$ matrix, $\texttt{x}\in \mathbb{L}_{k}^{n}, \textrm{M}^{\otimes^{k}}\texttt{x}=\textrm{M}^{\otimes^{k}}\left ( \texttt{x}\right )$, we have matrix associativity as: $\left (\textrm{M}'\textrm{M} \right ){\otimes^{k}}\texttt{x}=\texttt{M}'{\otimes^{k}}\left ( {\texttt{M}\otimes^{k}}\texttt{x}\right )$.

A crucial distinction between Lorentzian matrix-vector multiplication and other matrix-vector multiplications on the hyperboloid model lies in the size of the matrix $\texttt{M}$.
Assuming that a $n$-dimensional feature needs to be transformed into a $m$-dimensional feature, it is natural for the size of the matrix $\texttt{M}$ to be $m\times n$.
This requirement is fulfilled by Lorentzian matrix-vector multiplication. 
However, in other methods \citep{chami2019hyperbolic, liu2019hyperbolic}, the size of the matrix $\texttt{M}$ is $\left ( m+1\right )\times \left ( n+1\right )$, which leads to the constraint of tangent spaces cannot be satisfied. 
Therefore, the Lorentzian transformation strictly adheres to hyperbolic geometry and effectively preserves the graph structure and properties within the hyperbolic space. 

As shown in Figure \ref{fig2}, after acquiring $\texttt{H}^{\left ( n\right )}$, we apply the exponential mapping (Eq. \ref{equation1}) to map the learned node features $\texttt{H}^{\left ( n\right )}$ into the hyperbolic space and then perform Lorentzian linear transformation within the hyperbolic space using Eq. \ref{equation4}. 
This specific process is given by the following:
\begin{equation}
\small
    \textrm{M}^{\otimes^{k}}\left ( \texttt{H}^{\left ( n\right )}\right )=\text{exp}_{0}^{k}\left ( \hat{\textrm{M}}\left (\text{log}_{0}^{k}\left(\text{exp}_{0}^{k}\left (\texttt{H}^{\left ( n\right )}\right )\right )\right )\right ),
\end{equation}
where $\text{exp}_{0}^{k}: \mathcal{T} _{0}\mathbb{L}_{k}^{n}\rightarrow\mathbb{L}_{k}^{m}$, $\text{log}_{0}^{k}: \mathbb{L}_{k}^{n}\rightarrow\mathcal{T} _{0}\mathbb{L}_{k}^{m}$. 
$k$ is the constant curvature.
Since the tangent space $\mathcal{T}_{0}\mathbb{L}^{n}$ is a subspace of Euclidean space $\mathbb{R}^{n}$, Euclidean linear transformations can be applied to tangent spaces. 

\subsection{Graph Node Prediction}
Finally, we use logarithm mapping (Eq. \ref{equation2}) to map the transformed node features back into Euclidean space for node prediction. 
The process is as follows:
\begin{equation}
\begin{aligned}
    \texttt{H}^{\left ( n\right )'} &= \text{log}_{0}^{k}\left ( \textrm{M}^{\otimes^{k}}\left ( \texttt{H}^{\left ( n\right )}\right )\right ),\\
   \hat{\texttt{Y}} &= \text{argmax}\left ( \texttt{H}^{\left ( n\right )'}\right ),
\end{aligned}
\end{equation}
where $\hat{\texttt{Y}}$ is the index of the maximum probability in the category.

\renewcommand{\arraystretch}{1.2}

\section{Experiments} 
\subsection{Datasets}
To validate the effectiveness and robustness of our framework, we conduct experiments on five publicly node classification datasets.
The task can be performed under both semi-supervise and fully supervised learning.
We used Cora, Citeseer, and PubMed \citep{sen2008collective} with semi-supervised learning.
In these datasets, papers and their citation links are modeled as graphs, where nodes are papers and edges are citation links. 
For a fair comparison, we adopted the data splitting and evaluation method commonly used in previous studies.

With fully supervised learning, we use another two real world networks datasets, including Disease and Airport datasets \citep{zhang2021lorentzian}. 
The Disease dataset is a graph with tree structure where node features indicate the susceptibility to the disease. 
The Airport dataset is a dataset where nodes represent airports and edges are the airline routes. 
In the Airport dataset, we utilize one-hot encoding to represent the nodes as node features.
We split nodes in Disease dataset into $30/10/60\%$ and Airport dataset into $70/15/15\%$ for training, validation, and test sets, respectively. 
These split ratio is in line with \citep{chami2019hyperbolic, zhang2021lorentzian, chen2022fully}.

These five datasets statistics are described in Table \ref{tab:datastat}, where $\delta$ refers to Gromovs $\delta$-hyperbolicity.
The lower $\delta$, the more tree-like of the graph. 
From Table~\ref{tab:datastat}, we see that the Cora dataset has the highest $\delta$, which indicates that the hierarchical structure of Cora is not obvious.

\begin{table}[h]
\centering
\scriptsize
\begin{tabular}{cccccccc}
\hline
\textbf{Dataset} & \textbf{Nodes} & \textbf{Edges} & \textbf{Label} & \textbf{Features}  & $\bm{\delta}$     \\ 
\hline
Cora             & $2,708$           & $5,429 $          & $7$              & $1,433$    &$11$        \\
Citeseer         & $3,327$           & $4,732$           & $6$              & $3,703$    &$4.5$      \\
PubMed           & $19,717$          & $44,338$          & $3$              & $500$      &$3.4$       \\
Disease          & $1,044$           & $1,043$           & $2$              & $1,000$    &$0$         \\
Airport          & $3,188$           & $18,631$          & $4$              & $3,188$    &$1$          \\
\hline
\end{tabular}
\caption{\label{tab:datastat} Dataset statistic on five datasets.}
\end{table}

\begin{table*}[h!]
\centering
\begin{tabular}{cclccc}
\hline
\multicolumn{1}{c}{\textbf{Space}} &\textbf{Type} & \textbf{Method} & \makecell{\textbf{Cora} \\$\delta =11$} & \makecell{\textbf{Citeseer} \\$\delta =4.5$} & \makecell{\textbf{PubMed} \\$\delta =3.4$}     \\ 
\hline
\multirow{11}{*}{Euclidean}              &             & GCN\citep{kipf2017semi}              & $81.5 $                         & $70.3  $                             & $79 $                              \\
              &             & GAT\citep{velivckovic2018graph}      & $83.0$                          & $72.5\pm 0.7$                        & $79.0\pm 0.3$                      \\
              &Nonlinear   & APPNP\citep{gasteiger2018predict}    & $83.3 $                         & $71.8$                               & $80.1$                             \\
              &             & GraphHeat\citep{xu2019graph}         & $83.7$                          & $72.5\pm 0.7$                        & $80.5$                             \\
              &             & ElasticGNN\citep{liu2021elastic}     & $83.7\pm 0.2$                   & $72.2\pm 0.6 $                       & $80.5\pm 0.1$                      \\
              &             & SCGNN\citep{liu2023self}             & $\textbf{84.5}\pm \textbf{0.3}$ & $73.5\pm 0.5 $                       & $80.8\pm 0.5$                      \\
\cline{2-6}      
              &             & SGC\citep{wu2019simplifying}         & $81.0\pm 0.0 $                  & $71.9\pm 0.1$                        & $78.9\pm 0.0$                      \\
              &             & SIGN-linear\citep{frasca2020sign}    & $81.7$                          & $72.4$                               & $78.6$                             \\
              &Linear       & DGC\citep{wang2021dissecting}        & $83.3\pm 0.0$                   & $73.3\pm 0.1$                        & $80.3\pm 0.1$                      \\            
              &             & G$^{2}$CN\citep{li2022g}             & $82.7$                          & $73.8$                               & $80.4$                             \\
              &             & FLGC\citep{cai2023fully}             & $84.0\pm 0.0$                   & $73.2\pm 0.0$                        & $81.1\pm 0.0$                      \\
\hline
              &             & HGCN\citep{chami2019hyperbolic}      & $81.3\pm 0.6$                   & $70.9\pm 0.6$                        & $78.4\pm 0.4$                      \\
              &             & HAT\citep{zhang2021hyperbolic}       & $83.1\pm 0.6$                   & $71.9\pm 0.6$                        & $78.6\pm 0.5$                      \\
Hyperbolic    &Nonlinear   & LGCN\citep{zhang2021lorentzian}      & $83.3\pm 0.7$                   & $71.9\pm 0.7$                        & $78.6\pm 0.7$                       \\
              &             & HYBONET\citep{chen2022fully}         & $80.2\pm 1.3$                   & -                                    & $78.0\pm 1.0$                       \\ 
              &             & HGCL\citep{liu2022enhancing}         & $82.3\pm 0.5$                   & $72.1\pm 0.6$                        & $79.14\pm 0.7$                      \\
                   
\cline{2-6}
              &Linear       & \textbf{$\text{L}^{2}$GC}(ours)                  & $82.4\pm 0.0$                   & $\textbf{74.7}\pm \textbf{0.0}$      & $\textbf{81.3}\pm \textbf{0.0}$      \\
\hline 
\end{tabular}
\caption{\label{tab:mainresult} Test accuracy ($\%$) of semi-supervised node classification on citation networks. "-" indicates that there is no result reported for this dataset in the literature.}
\end{table*}

\subsection{Baselines and Setup} 
\textbf{Baselines.} In the evaluation on citation network datasets, we compared our model $\text{L}^{2}$GC with 16 baselines.
These baselines are grouped into four categories by their space used and type of GCN.
\begin{itemize}
    \item {Nonlinear models with Euclidean space. These models are traditional GCN and their variants, they are GCN \citep{kipf2017semi}, GAT \citep{velivckovic2018graph}, APPNP \citep{gasteiger2018predict}, GraphHeat \citep{xu2019graph}, ElasticGNN \citep{liu2021elastic}, and SCGNN \citep{liu2023self}.}
    \item {Linear models with Euclidean space. These models used liner transformation in GCN, they are SGC \citep{wu2019simplifying}, SIGN-linear \citep{frasca2020sign}, DGC \citep{wang2021dissecting}, G$^{2}$CN \citep{li2022g} and FLGC \citep{cai2023fully}.}
    \item {Nonlinear models with hyperbolic space. These model extend GCN to hyperbolic space, they are HGCN \citep{chami2019hyperbolic}, HAT \citep{zhang2021hyperbolic}, LGCN \citep{zhang2021lorentzian}, HGCL \citep{liu2022enhancing} and HYBONET \citep{chen2022fully}.}
    \item {Linear model with hyperbolic space. Our model ($\text{L}^{2}$GC) extend linear GCN to hyperbolic space. It is worth noting that there are no other linear models on hyperbolic spaces. Our proposed $\text{L}^{2}$GC is the first simplification of nonlinear GCN models in hyperbolic spaces. }
\end{itemize}
For a fair comparison, we report the best results of these models from the corresponding literature. 
Due to space limitations, a detailed description of these baselines is presented in Appendix \ref{app}.

\textbf{Setup.} To train our model, we employ the Adam optimizer \citep{kingma2014adam} and take cross-entropy function as our loss function. 
The hyperparameters are determined by Bayesian optimization. 
The parameter optimization scopes are: learning rate ($\lambda$): $\left [ 0,2\right ]$;  weight decay ($w$): $\left [ 1^{-10},1^{-1}\right ]$; retention probability ($\alpha$): $\left [ 0,1\right ]$;  propagation step ($n$): $\left [ 0,30\right ]$. 
For the bias $b$, $1$ represents the utilization of bias and $0$ denotes the opposite. 
The dropout probability $d$ is set to $0$ for all experiments. 
The curvature $K$ of the hyperbolic space is 1.

Finally, the hyperparameters used in our experiments are given in Table \ref{table3}. 
All experiments were conducted on a machine with an NVIDIA GeForce GTX 1660 GPU with 6GB memory. 
We evaluate our model using the accuracy rate (ACC). 
The results are obtained by averaging the results of ten random runs. 
\begin{table}[h]
\centering
\begin{tabular}{cccccc}
\hline
\text{Dataset}   & \text{$\lambda$} & \text{$w$}        & \text{$\alpha$}   & \text{$n$} & \text{$b$}      \\ 
\hline
Cora             & $0.6$            & $3.0^{-5} $             & $0.1$              & $20$       & $1$         \\
Citeseer         & $1.2$            & $9.8^{-5} $            & $0.1$              & $20$       & $1$         \\
PubMed           & $0.94$           & $8.8^{-6} $             & $0.05$             & $23$       & $0$        \\
Disease          & $1.6$            & $1.9^{-8} $             & $0.1$              & $4$        & $1$        \\
Airport          & $1.0$            & $1.1^{-10} $            & $0.1$              & $9$        & $0$         \\
\hline
\end{tabular}
\caption{\label{table3} Hyper-parameter settings of our model.}
\end{table}

\subsection{Performance on Semi-supervised Node Classification}
We present a summary of experimental results in Table \ref{tab:mainresult}.
We analyse the results as follow:

\noindent
\textbf{Comparing with models in Euclidean spaces.} 
Our $\text{L}^{2}$GC outperforms all previous SOTA nonlinear and linear models on both Citeseer and PubMed datasets.
These datasets exhibit a distinct tree-like structure characterized by smaller values of $\delta$. 
We attribute these successes to the ability of hyperbolic spaces to effectively capture the hierarchical structure implicitly in the data. 
Specifically, compared to SGC, our $\text{L}^{2}$GC achieves improvements by 1.2$\%$, 3.8$\%$ and 3.0$\%$  absolutely on Cora, Citeseer, and PubMed, respectively. 
Compared with the current SOTA linear model (G$^{2}$CN), our approach achieves improvement by 1.2$\%$ and 0.2$\%$  absolutely on the Citeseer and PubMed. 
However, we observed that our approach does not exhibit a significant advantage when compared to most models on Cora dataset, which has the biggest $\delta$. 
We think that the absence of a tree-like structure in the Cora dataset might be the reason why our approach does not perform as well on this task. 

\noindent
\textbf{Comparing with models in hyperbolic spaces.} 
Compared to nonlinear models with hyperbolic space, our approach also demonstrate superior performance on the Citeseer and PubMed datasets, since our model performs Lorentzian linear transformation operations solely in hyperbolic space. 
Specifically, compared with the recent SOTA model (HGCL), our approach has improved by 3.6$\%$ and 2.7$\%$ on Citeseer and PubMed datasets, respectively. 
However, our model fail behind the models in Euclidean space. 
This observation in line with other previous hyperbolic models, like~\citep{chami2019hyperbolic, zhang2021hyperbolic, zhang2021lorentzian, liu2022enhancing, chen2022fully}.
It is worth noting that there is no GCN model of linear type in hyperbolic space. 
Our proposed approach represents a notable advance, as it is the first attempt to simplify nonlinear GCN models specifically designed for hyperbolic spaces. 
This highlights the unique contribution and advancement of our proposed approach.

\begin{table}[h]
\centering
\small
\begin{tabular}{clcc}
\hline 
\textbf{Space} & \textbf{Method} & \makecell{\textbf{Disease}  \\ $\delta =0$}  & \makecell{\textbf{Airport} \\$\delta =1$}       \\ 
\hline
\multirow{4}{*}{$\mathbb{E}$}   &GCN\citeyearpar{kipf2017semi}       & $69.7\pm 0.4 $    & $81.4\pm 0.6 $           \\
                                &GAT\citeyearpar{velivckovic2018graph}   & $70.4\pm 0.4 $     & $81.5\pm 0.3 $    \\
                                &SGC\citeyearpar{wu2019simplifying}      & $69.5\pm 0.2 $     & $80.6\pm 0.1 $     \\
                                &SCGNN\citeyearpar{liu2023self}        & $85.3 \pm 0.4$       & -              \\
\hline
\multirow{6}{*}{$\mathbb{H}$}       &HGCN\citeyearpar{chami2019hyperbolic}     & $82.8\pm 0.8 $      & $90.6\pm 0.2 $      \\
                 &HAT\citeyearpar{zhang2021hyperbolic}     & $83.6\pm 0.9 $                  & -                     \\
                 &LGCN\citeyearpar{zhang2021lorentzian}     & $84.4\pm 0.8 $                  & $90.9\pm 1.7 $          \\
                 &HGCL\citeyearpar{liu2022enhancing}        & $93.4\pm 0.8 $                  & $92.3\pm 1.0 $         \\
                 &HYBONET\citeyearpar{chen2022fully}        & $\textbf{96.0} \pm \textbf{1.0}$          & $90.9\pm 1.4 $          \\
                 &\textbf{$\text{L}^{2}$GC}(ours)           & $94.4\pm 0.1$ & $\textbf{94.0}\pm \textbf{0.1}$   \\
\hline 
\end{tabular}
\caption{\label{tab:fullsupresult} Test accuracy ($\%$) of fully supervised node classification on Disease and Airport datasets. $\mathbb{E}$ and $\mathbb{H}$ represent Euclidean space and hyperbolic space, respectively. "-" indicates that there is no report for this dataset in the literature.}
\end{table}

\subsection{Performance on Fully supervised Node Classification}
We further validate the effectiveness and robustness of our approach in different domains under fully supervised node classification.
We utilize Disease and Airport two datasets, which have lower $\delta$ values. 
Other settings in these experiments remain consistent with semi-supervised tasks.
Here, we compare with GCN, GAT, HGCN, SGC, HAT, LGCN, HGCL, HYBONET, and SCGNN models. 
For a fair comparison, we provide the results reported in the corresponding literature.
The results are shown in Table \ref{tab:fullsupresult}. 

In both datasets, our model improved large margin over Euclidean space model since the datasets have lower $\delta$ value.
In hyperbolic space, our model achieves the best accuracy with the Airport dataset.
Our model improves 3.4$\%$ and 1.8$\%$ over HYBONET and HGCL on accuracy, respectively. 
On the Disease dataset, our approach demonstrates an improvement over most previous models. 
Although $\text{L}^{2}$GC does not show a significant improvement over HYBONET on the Disease dataset, our model exhibits advantages in terms of reducing training time and increasing stability.
Ultimately, this result highlights the ability of our model to effectively capture the hierarchical structure present in the data. 
Our model further enhances the performance of linear GCNs and demonstrates the advantage of leveraging hyperbolic geometry. 

\begin{table}[h]
\centering
\begin{tabular}{lccc}
\hline
  &  \makecell{\textbf{Cora} \\$\delta =11$} & \makecell{\textbf{Citeseer} \\$\delta =4.5$} & \makecell{\textbf{PubMed} \\$\delta =3.4$}    \\ 
\hline
    $Variant\;I$        & $\textbf{83.1}$           & $73.0 $              & $79.4 $   \\
    $Variant\;II$       & $80.2 $                   & $73.6 $              & $79.7 $      \\
    $\textbf{L}^{2}\textbf{GC}$   & $82.4 $                   & $\textbf{74.7}$      & $\textbf{81.3} $   \\
\hline 
\end{tabular}
\caption{\label{tab:abla} The ablation experiments of $\text{L}^{2}$GC.} 
\end{table}

\subsection{Analysis and Visualization}
To analyse the effective and efficiency of our model, we further do extensive experiments and visualize some classification results.

\textbf{Ablations.} To analyse the impact of the components in $\text{L}^{2}\text{GC}$, we explore the performance of two model variants. 
\begin{itemize}
    \item {$Variant\;I:$ $\text{L}^{2}\text{GC}$ with Personalized Propagation scheme and without Lorentz model, which indicates the features transformation stage in Euclidean space.}
    \item {$Variant\;II:$ $\text{L}^{2}\text{GC}$ without Personalized Propagation scheme and with Lorentz model, we use the SGC propagation scheme in the feature propagation stage.}
\end{itemize}
We compare the test accuracy of all variants against $\text{L}^{2}\text{GC}$ in Table \ref{tab:abla}. 
It is clear that the $\text{L}^{2}\text{GC}$ outperforms the other two variants on Citeseer and PubMed datasets. 
However, on the Cora dataset, we observe that the performance of $Variant\;I$ improves without the hyperbolic component.
This result could be attributed to the largest $\delta$ of the dataset. 
This observation demonstrates that hyperbolic space is suitable for tree-like structure.

\begin{figure}[h!]
\begin{center}
\includegraphics[width=1.1\linewidth]{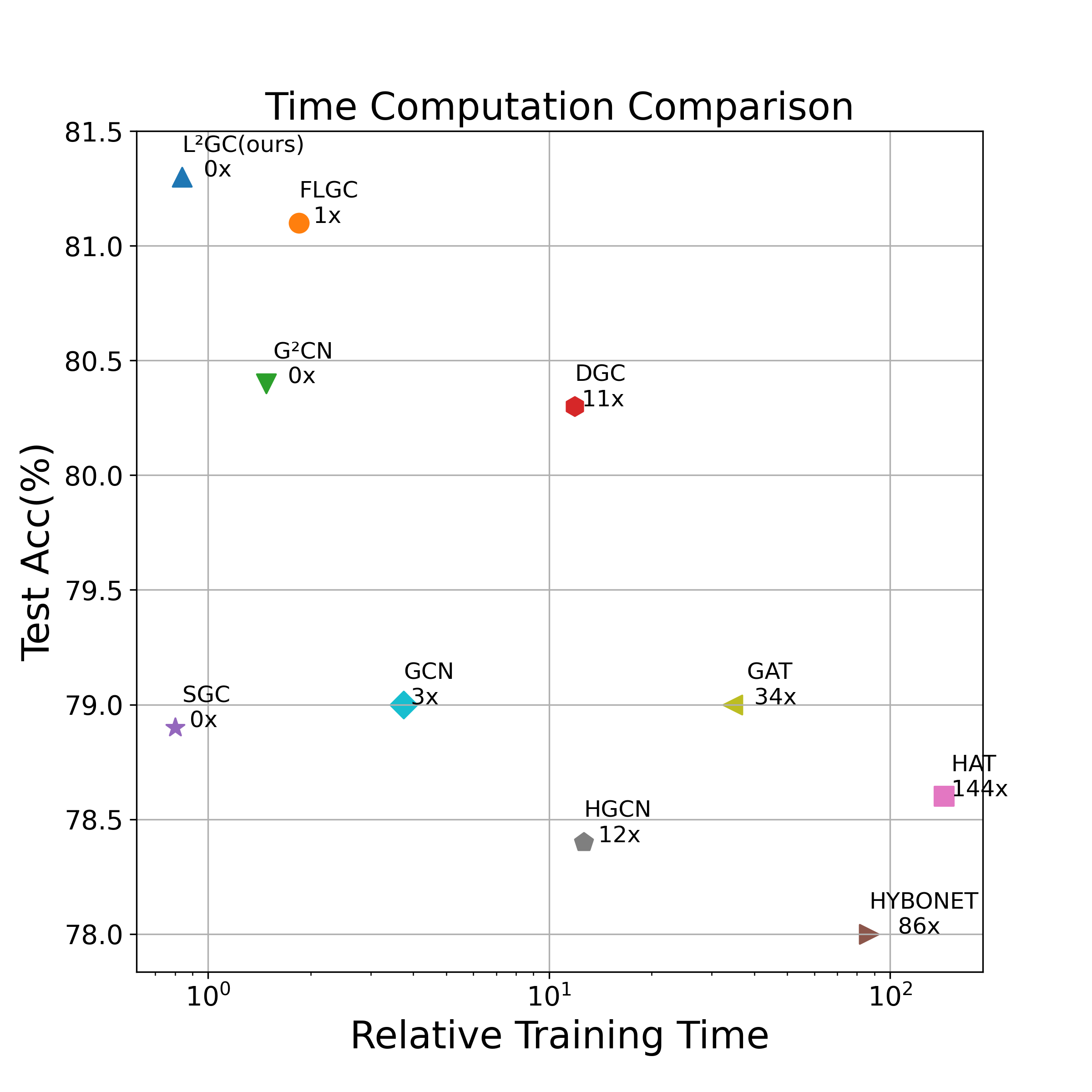} 
\caption{Comparison over training time on PubMed dataset.}
\label{fig:time}
\end{center}
\end{figure}

\textbf{Efficiency.}
To further validate the efficiency of our model, we compared its performance with other models, such as GCN, GAT, HGCN, HAT, HYBONET, SGC, DGC, G$^2$CN and FLGC. 
We plot the performance of these models relative to the training time of $\text{L}^{2}$GC on the PubMed dataset in Figure \ref{fig:time}.
The training time of $\text{L}^{2}$GC includes the feature propagation time for a fair comparison. 
Overall, Figure \ref{fig:time} shows that $\text{L}^{2}$GC is trained faster one or two orders of magnitude than other nonlinear models.
We measure the training time on the same NVIDIA GeForce GTX 1660 GPU with 6GB memory. 

Moreover, our model also simplifies the hyperbolic model by leveraging the structure of the linear model.
We compared the amount of parameters with HYBONET, as shown in Table \ref{tab:paranum}.
It obviously shows that our model can maintain superior performance with significantly fewer parameters. 

\begin{table}[h]
\centering
\begin{tabular}{lcc}
\hline
\textbf{Model}  &  HYBONET  & $\text{L}^{2}$GC(ours)     \\ 
\hline
    Cora                & $23356$       & $8604$ ($\downarrow$ $63.16\%$)  \\
    Citeseer            & $59659$       & $18520$ ($\downarrow$ $68.95\%$ ) \\
    PubMed              & $8360$        & $1002$  ($\downarrow$ $88.01\%$)  \\
\hline 
\end{tabular}
\caption{\label{tab:paranum} Comparison on parameters number.} 
\end{table}

\begin{figure}[h!]
\begin{center}
\includegraphics[width=1\linewidth]{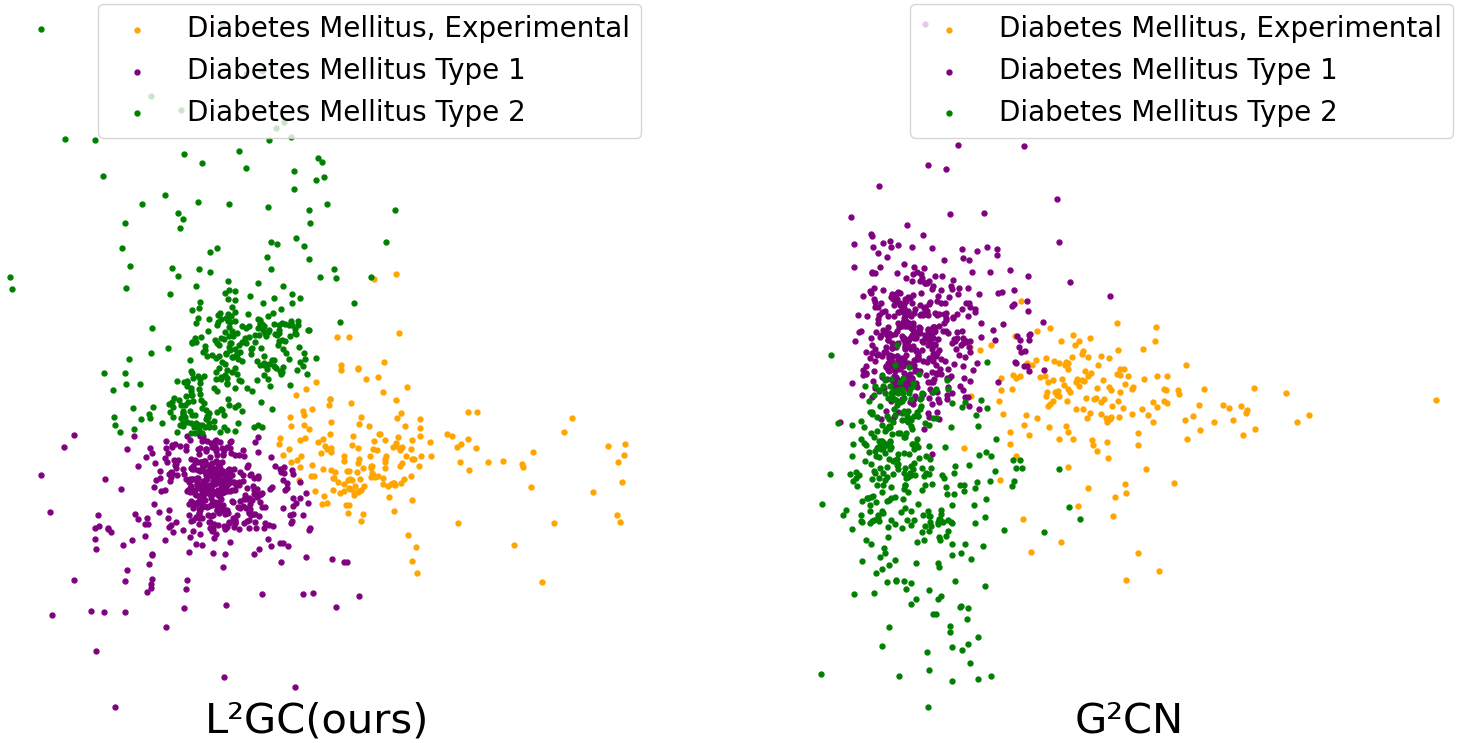} 
\caption{Comparison node classification result with linear model ($\text{G}^{2}$CN). The color indicates the label category of nodes.}
\label{fig4}
\end{center}
\end{figure}

\begin{figure}[h!]
\begin{center}
\includegraphics[width=1\linewidth]{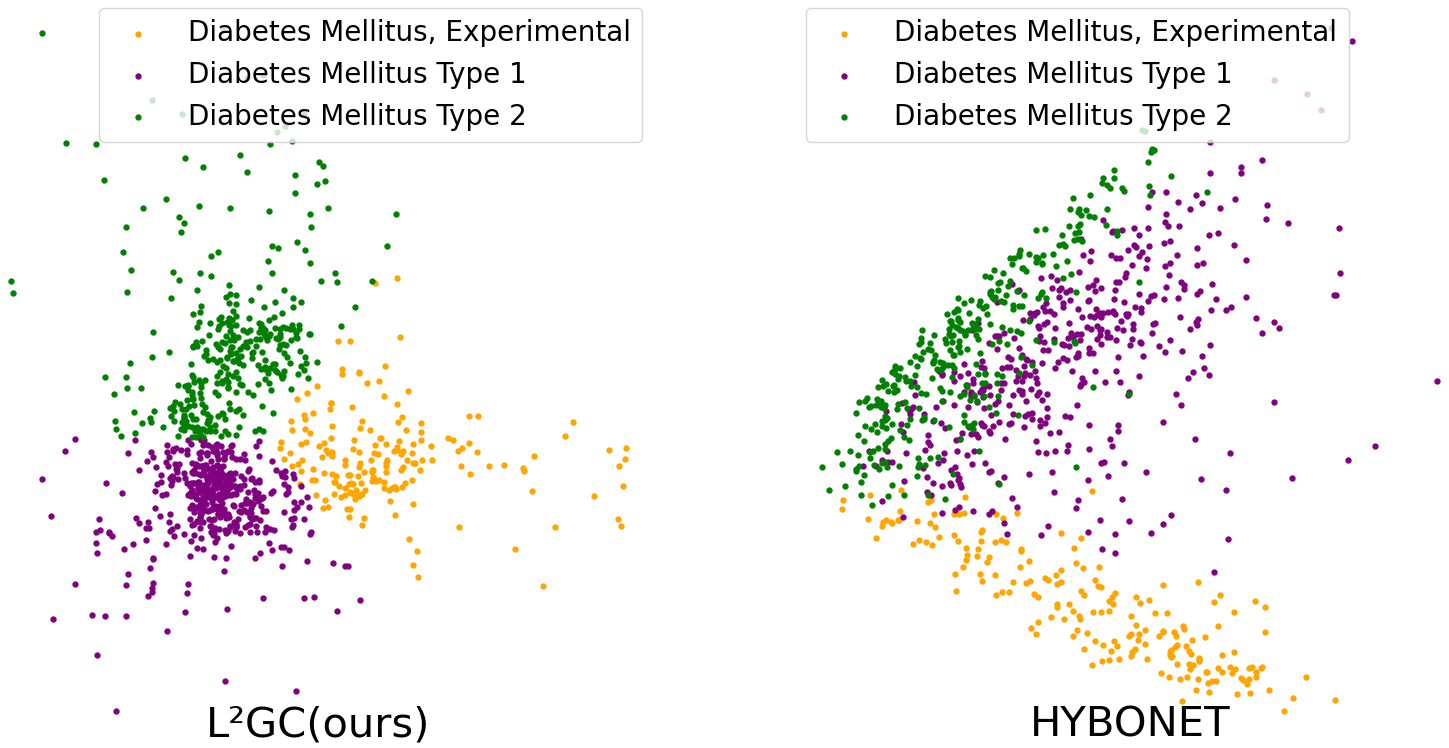} 
\caption{Comparison node classification result with hyperbolic model (HYBONET). The color indicates the label category of nodes.}
\label{fig5}
\end{center}
\end{figure}

\textbf{Visualizations.} 
We visualized the test set of PubMed dataset and its predicted label distribution in two dimensions with Principal Component Analysis (PCA). 
We compared the classification results with $\text{G}^{2}$CN and HYBONET on PubMed dataset.
The results are shown in Figure \ref{fig4} and Figure \ref{fig5}, respectively.
From these figures, we see that our model can achieve better node classification results.

\section{Conclusion and Future Work}
In this paper, we propose a novel Lorentzian Linear Graph Convolutional Networks framework for node classification based on hyperbolic space. 
Our work is the first generalization of the linear GCN model to hyperbolic space, which capturing of the hierarchical structure in the data.
Our approach not only leverages the strengths of the linear model, but also integrates the properties of hyperbolic spaces to achieve new SOTA results on the Citeseer and PubMed datasets.
However, there are limitations to our approach, when compared to other Euclidean space models with large $\delta$ value. 
In the future, we plan to explore modeling our approach in a mixed spaces to enhance the ability of our model.

\section*{Acknowledgements}
This work is supported by National Natural Science Foundation of China (Nos.62066033, 61966025);
Inner Mongolia Natural Science Foundation (Nos.2020BS06001, 2022JQ05); 
Inner Mongolia Autonomous Region Science and Technology Programme Project (Nos.2023YFSW0001, 2022YFDZ0059);
Collaborative Innovation Project of Universities and Institutes in Hohhot City.
We also thank Qing Zhang for his helpful discussions.



\section*{References}
\bibliographystyle{lrec-coling2024-natbib}
\bibliography{lrec-coling2024-example}

\begin{thebibliography}{34}
\expandafter\ifx\csname natexlab\endcsname\relax\def\natexlab#1{#1}\fi

\bibitem[{Cai et~al.(2023)Cai, Zhang, Ghamisi, Cai, Liu, and Ding}]{cai2023fully}
Yaoming Cai, Zijia Zhang, Pedram Ghamisi, Zhihua Cai, Xiaobo Liu, and Yao Ding. 2023.
\newblock Fully linear graph convolutional networks for semi-supervised and unsupervised classification.
\newblock \emph{ACM Transactions on Intelligent Systems and Technology}, 14(3):1--23.

\bibitem[{Chami et~al.(2019)Chami, Ying, R{\'e}, and Leskovec}]{chami2019hyperbolic}
Ines Chami, Rex Ying, Christopher R{\'e}, and Jure Leskovec. 2019.
\newblock Hyperbolic graph convolutional neural networks.
\newblock \emph{Advances in neural information processing systems}, 32:4869--4880.

\bibitem[{Chen et~al.(2022)Chen, Han, Lin, Zhao, Liu, Li, Sun, and Zhou}]{chen2022fully}
Weize Chen, Xu~Han, Yankai Lin, Hexu Zhao, Zhiyuan Liu, Peng Li, Maosong Sun, and Jie Zhou. 2022.
\newblock Fully hyperbolic neural networks.
\newblock In \emph{Proceedings of the 60th Annual Meeting of the Association for Computational Linguistics (Volume 1: Long Papers)}, pages 5672--5686.

\bibitem[{Clauset et~al.(2008)Clauset, Moore, and Newman}]{clauset2008hierarchical}
Aaron Clauset, Cristopher Moore, and Mark~EJ Newman. 2008.
\newblock Hierarchical structure and the prediction of missing links in networks.
\newblock \emph{Nature}, 453(7191):98--101.

\bibitem[{Feng et~al.(2022)Feng, Cai, Wei, and Li}]{feng2022graph}
Lixia Feng, Yongqi Cai, Erling Wei, and Jianwu Li. 2022.
\newblock Graph neural networks with global noise filtering for session-based recommendation.
\newblock \emph{Neurocomputing}, 472:113--123.

\bibitem[{Frasca et~al.(2020)Frasca, Rossi, Eynard, Chamberlain, Bronstein, and Monti}]{frasca2020sign}
Fabrizio Frasca, Emanuele Rossi, Davide Eynard, Ben Chamberlain, Michael Bronstein, and Federico Monti. 2020.
\newblock Sign: Scalable inception graph neural networks.
\newblock \emph{arXiv preprint arXiv:2004.11198}.

\bibitem[{Ganea et~al.(2018)Ganea, B{\'e}cigneul, and Hofmann}]{ganea2018hyperbolic}
Octavian Ganea, Gary B{\'e}cigneul, and Thomas Hofmann. 2018.
\newblock Hyperbolic neural networks.
\newblock In \emph{Advances in neural information processing systems}, pages 5345--5355.

\bibitem[{Gasteiger et~al.(2018)Gasteiger, Bojchevski, and G{\"u}nnemann}]{gasteiger2018predict}
Johannes Gasteiger, Aleksandar Bojchevski, and Stephan G{\"u}nnemann. 2018.
\newblock Predict then propagate: Graph neural networks meet personalized pagerank.
\newblock In \emph{International Conference on Learning Representations (ICLR)}.

\bibitem[{Gulcehre et~al.(2018)Gulcehre, Denil, Malinowski, Razavi, Pascanu, Hermann, Battaglia, Bapst, Raposo, Santoro et~al.}]{gulcehre2018hyperbolic}
Caglar Gulcehre, Misha Denil, Mateusz Malinowski, Ali Razavi, Razvan Pascanu, Karl~Moritz Hermann, Peter Battaglia, Victor Bapst, David Raposo, Adam Santoro, et~al. 2018.
\newblock Hyperbolic attention networks.
\newblock \emph{arXiv preprint arXiv:1805.09786}.

\bibitem[{Kingma and Ba(2014)}]{kingma2014adam}
Diederik~P Kingma and Jimmy Ba. 2014.
\newblock Adam: A method for stochastic optimization.
\newblock \emph{arXiv preprint arXiv:1412.6980}.

\bibitem[{Kipf and Welling(2017)}]{kipf2017semi}
Thomas~N. Kipf and Max Welling. 2017.
\newblock Semi-supervised classification with graph convolutional networks.
\newblock In \emph{International Conference on Learning Representations (ICLR)}.

\bibitem[{Li and Goldwasser(2019)}]{li2019encoding}
Chang Li and Dan Goldwasser. 2019.
\newblock Encoding social information with graph convolutional networks forpolitical perspective detection in news media.
\newblock In \emph{Proceedings of the 57th Annual Meeting of the Association for Computational Linguistics}, pages 2594--2604.

\bibitem[{Li et~al.(2022)Li, Guo, Wang, Wang, and Lin}]{li2022g}
Mingjie Li, Xiaojun Guo, Yifei Wang, Yisen Wang, and Zhouchen Lin. 2022.
\newblock G$^{2}${CN}: Graph gaussian convolution networks with concentrated graph filters.
\newblock In \emph{International Conference on Machine Learning}, pages 12782--12796. PMLR.

\bibitem[{Liu et~al.(2022)Liu, Yang, Zhou, Feng, and Fournier-Viger}]{liu2022enhancing}
Jiahong Liu, Menglin Yang, Min Zhou, Shanshan Feng, and Philippe Fournier-Viger. 2022.
\newblock Enhancing hyperbolic graph embeddings via contrastive learning.
\newblock \emph{arXiv preprint arXiv:2201.08554}.

\bibitem[{Liu et~al.(2019)Liu, Nickel, and Kiela}]{liu2019hyperbolic}
Qi~Liu, Maximilian Nickel, and Douwe Kiela. 2019.
\newblock Hyperbolic graph neural networks.
\newblock \emph{Advances in neural information processing systems}, 32:12--22.

\bibitem[{Liu et~al.(2021)Liu, Jin, Ma, Li, Liu, Wang, Yan, and Tang}]{liu2021elastic}
Xiaorui Liu, Wei Jin, Yao Ma, Yaxin Li, Hua Liu, Yiqi Wang, Ming Yan, and Jiliang Tang. 2021.
\newblock Elastic graph neural networks.
\newblock In \emph{International Conference on Machine Learning}, pages 6837--6849. PMLR.

\bibitem[{Liu et~al.(2023)Liu, Zhao, Wang, Geng, Xiao, and Lin}]{liu2023self}
Yanbei Liu, Shichuan Zhao, Xiao Wang, Lei Geng, Zhitao Xiao, and Jerry Chun-Wei Lin. 2023.
\newblock Self-consistent graph neural networks for semi-supervised node classification.
\newblock \emph{IEEE Transactions on Big Data}, 9(4):1186--1197.

\bibitem[{Muscoloni et~al.(2017)Muscoloni, Thomas, Ciucci, Bianconi, and Cannistraci}]{muscoloni2017machine}
Alessandro Muscoloni, Josephine~Maria Thomas, Sara Ciucci, Ginestra Bianconi, and Carlo~Vittorio Cannistraci. 2017.
\newblock Machine learning meets complex networks via coalescent embedding in the hyperbolic space.
\newblock \emph{Nature communications}, 8(1):1615.

\bibitem[{Nickel and Kiela(2018)}]{nickel2018learning}
Maximillian Nickel and Douwe Kiela. 2018.
\newblock Learning continuous hierarchies in the lorentz model of hyperbolic geometry.
\newblock In \emph{International conference on machine learning}, pages 3779--3788. PMLR.

\bibitem[{Sankar et~al.(2021)Sankar, Liu, Yu, and Shah}]{sankar2021graph}
Aravind Sankar, Yozen Liu, Jun Yu, and Neil Shah. 2021.
\newblock Graph neural networks for friend ranking in large-scale social platforms.
\newblock In \emph{Proceedings of the Web Conference 2021}, pages 2535--2546.

\bibitem[{Sen et~al.(2008)Sen, Namata, Bilgic, Getoor, Galligher, and Eliassi-Rad}]{sen2008collective}
Prithviraj Sen, Galileo Namata, Mustafa Bilgic, Lise Getoor, Brian Galligher, and Tina Eliassi-Rad. 2008.
\newblock Collective classification in network data.
\newblock \emph{AI magazine}, 29(3):93--93.

\bibitem[{Shang et~al.(2019)Shang, Xiao, Ma, Li, and Sun}]{shang2019gamenet}
Junyuan Shang, Cao Xiao, Tengfei Ma, Hongyan Li, and Jimeng Sun. 2019.
\newblock Gamenet: Graph augmented memory networks for recommending medication combination.
\newblock In \emph{Proceedings of the AAAI conference on artificial intelligence}, volume~33, pages 1126--1133.

\bibitem[{Tian et~al.(2021)Tian, Chen, Song, and Wan}]{tian2021dependency}
Yuanhe Tian, Guimin Chen, Yan Song, and Xiang Wan. 2021.
\newblock Dependency-driven relation extraction with attentive graph convolutional networks.
\newblock In \emph{Proceedings of the 59th Annual Meeting of the Association for Computational Linguistics and the 11th International Joint Conference on Natural Language Processing (Volume 1: Long Papers)}, pages 4458--4471.

\bibitem[{Tifrea et~al.(2018)Tifrea, B{\'e}cigneul, and Ganea}]{tifrea2018poincar}
Alexandru Tifrea, Gary B{\'e}cigneul, and Octavian-Eugen Ganea. 2018.
\newblock Poincar{\'e} glove: Hyperbolic word embeddings.
\newblock \emph{ArXiv}, abs/1810.06546.

\bibitem[{Veli{\v{c}}kovi{\'c} et~al.(2018)Veli{\v{c}}kovi{\'c}, Cucurull, Casanova, Romero, Li{\`o}, and Bengio}]{velivckovic2018graph}
Petar Veli{\v{c}}kovi{\'c}, Guillem Cucurull, Arantxa Casanova, Adriana Romero, Pietro Li{\`o}, and Yoshua Bengio. 2018.
\newblock Graph attention networks.
\newblock In \emph{International Conference on Learning Representations (ICLR)}.

\bibitem[{Wang et~al.(2021{\natexlab{a}})Wang, Chen, Pi, Yue, Wang, and Xu}]{wang2021self}
Yanda Wang, Weitong Chen, Dechang Pi, Lin Yue, Sen Wang, and Miao Xu. 2021{\natexlab{a}}.
\newblock Self-supervised adversarial distribution regularization for medication recommendation.
\newblock In \emph{IJCAI}, pages 3134--3140.

\bibitem[{Wang et~al.(2021{\natexlab{b}})Wang, Wang, Yang, and Lin}]{wang2021dissecting}
Yifei Wang, Yisen Wang, Jiansheng Yang, and Zhouchen Lin. 2021{\natexlab{b}}.
\newblock Dissecting the diffusion process in linear graph convolutional networks.
\newblock \emph{Advances in neural information processing systems}, 34:5758--5769.

\bibitem[{Willmore(2013)}]{willmore2013introduction}
Thomas~James Willmore. 2013.
\newblock \emph{An introduction to differential geometry}.
\newblock Courier Corporation.

\bibitem[{Wu et~al.(2019)Wu, Souza, Zhang, Fifty, Yu, and Weinberger}]{wu2019simplifying}
Felix Wu, Amauri Souza, Tianyi Zhang, Christopher Fifty, Tao Yu, and Kilian Weinberger. 2019.
\newblock Simplifying graph convolutional networks.
\newblock In \emph{International conference on machine learning}, pages 6861--6871. PMLR.

\bibitem[{Xu et~al.(2019)Xu, Shen, Cao, Cen, and Cheng}]{xu2019graph}
Bingbing Xu, Huawei Shen, Qi~Cao, Keting Cen, and Xueqi Cheng. 2019.
\newblock Graph convolutional networks using heat kernel for semi-supervised learning.
\newblock In \emph{Proceedings of the 28th International Joint Conference on Artificial Intelligence}, pages 1928--1934.

\bibitem[{Yao et~al.(2019)Yao, Mao, and Luo}]{yao2019graph}
Liang Yao, Chengsheng Mao, and Yuan Luo. 2019.
\newblock Graph convolutional networks for text classification.
\newblock In \emph{Proceedings of the AAAI conference on artificial intelligence}, volume~33, pages 7370--7377.

\bibitem[{Ying et~al.(2018)Ying, He, Chen, Eksombatchai, Hamilton, and Leskovec}]{ying2018graph}
Rex Ying, Ruining He, Kaifeng Chen, Pong Eksombatchai, William~L Hamilton, and Jure Leskovec. 2018.
\newblock Graph convolutional neural networks for web-scale recommender systems.
\newblock In \emph{Proceedings of the 24th ACM SIGKDD international conference on knowledge discovery \& data mining}, pages 974--983.

\bibitem[{Zhang et~al.(2021{\natexlab{a}})Zhang, Wang, Shi, Jiang, and Ye}]{zhang2021hyperbolic}
Yiding Zhang, Xiao Wang, Chuan Shi, Xunqiang Jiang, and Yanfang Ye. 2021{\natexlab{a}}.
\newblock Hyperbolic graph attention network.
\newblock \emph{IEEE Transactions on Big Data}, 8(6):1690--1701.

\bibitem[{Zhang et~al.(2021{\natexlab{b}})Zhang, Wang, Shi, Liu, and Song}]{zhang2021lorentzian}
Yiding Zhang, Xiao Wang, Chuan Shi, Nian Liu, and Guojie Song. 2021{\natexlab{b}}.
\newblock Lorentzian graph convolutional networks.
\newblock In \emph{Proceedings of the Web Conference 2021}, pages 1249--1261.

\end{thebibliography}

\appendix
\clearpage
\section{Appendix}
\label{app}
\textbf{Details of the baselines:}
Nonlinear models based on Euclidean space:
\begin{itemize}
    \item{\textbf{GCN}\citep{kipf2017semi} is the extension of convolutional neural networks to handle graph-structured data, which can obtain a better data representation.} 
    \item{\textbf{GAT}\citep{velivckovic2018graph} solves the problems of previous models based on graph convolution using masked-self attention layers.} 
    \item{\textbf{APPNP}\citep{gasteiger2018predict} uses the relationship between GCN and PageRank algorithm to derive an improved propagation scheme based on personalized PageRank.} 
    \item{\textbf{GraphHeat}\citep{xu2019graph} utilizes thermal kernels to enhance low-frequency filters and enhance the smoothness of signal changes on the graph.} 
    \item{\textbf{ElasticGNN}\citep{liu2021elastic} designs an Elastic GNN with a new message passing mechanism by adding $L_{1}$ regularization on top of traditional GNN's $L_{2}$ regularization.} 
    \item{\textbf{SCGNN}\citep{liu2023self} performs graph data augmentation and leverages a self-consistent constraint to maximize the mutual information of the unlabeled nodes across different augmented graph views.} 
\end{itemize}
Linear models based on Euclidean space:
\begin{itemize}
    \item{\textbf{SGC}\citep{wu2019simplifying} simplifies the GCN by removing the nonlinear activation function and reducing the entire GCN architecture to a direct multi-class logistic regression on preprocessed features.} 
    \item{\textbf{SIGN-linear}\citep{frasca2020sign} proposes a universal probability graph sampler that constructs training batches by sampling the original graph. The graph sampler can be designed according to different schemes.} 
    \item{\textbf{DGC}\citep{wang2021dissecting} decouples the terminal time and the feature propagation steps, making it more flexible and capable of exploiting a very large number of feature propagation steps.} 
    \item{\textbf{G}$^{2}$\textbf{CN}\citep{li2022g} has developed a new spectral analysis framework from a detailed spectral analysis called concentration analysis.}
    \item{\textbf{FLGC}\citep{cai2023fully} linearizes GCN and then decouple it into neighborhood propagation and prediction stages, resulting in a flexible framework.} 
\end{itemize}

Nonlinear models based on hyperbolic space:
\begin{itemize}
    \item{\textbf{HGCN}\citep{chami2019hyperbolic} aggregates the expressive power of GCN and hyperbolic geometry to learn node representations in scale-free graphs or hierarchical graphs} 
    \item{\textbf{HAT}\citep{zhang2021hyperbolic} introduces hyperbolic attention networks to endow neural networks with enough capacity to match the complexity of data with hierarchical and power-law structure.} \item{\textbf{LGCN}\citep{zhang2021lorentzian} reconstructs the Lorentz version of GCN, making the operation of the hyperbolic neural network strictly comply with the definition of hyperbolic geometry.} 
    \item{\textbf{HGCL}\citep{liu2022enhancing} takes advantage of contrastive learning to enhance the representation ability of the hyperbolic graph model.}
    \item{\textbf{HYBONET}\citep{chen2022fully} builds hyperbolic networks based on the Lorentz model by adapting the Lorentz transformations (including boost and rotation) to formalize essential operations of neural networks.} 
\end{itemize}

\end{document}